\documentclass[10pt,twocolumn,letterpaper]{article}
\PassOptionsToPackage{table}{xcolor}
\usepackage[pagenumbers]{iccv} 
\usepackage[table]{xcolor}
\definecolor{mygreen}{RGB}{0, 180, 0} 
\definecolor{myred}{RGB}{200,0,0}
\usepackage[accsupp]{axessibility}
\usepackage{amsmath}
\usepackage{algorithm}
\usepackage{algpseudocode}
\usepackage{graphicx} 

\usepackage{tcolorbox}
\usepackage{adjustbox}
\usepackage{courier} 
\usepackage{graphicx}
\usepackage{amsmath}
\usepackage{amssymb}
\usepackage{booktabs}
\usepackage{times}
\usepackage{epsfig}
\usepackage{graphicx}
\usepackage{booktabs} 
\usepackage{multirow} 

\definecolor{iccvblue}{rgb}{0.21,0.49,0.74}
\usepackage[pagebackref,breaklinks,colorlinks,allcolors=iccvblue]{hyperref}
\usepackage[capitalize]{cleveref}

\usepackage[accsupp]{axessibility} 

\def\paperID{18} 
\def\confName{ICCV}
\def\confYear{2025}

\title{LaViPlan : Language-Guided Visual Path Planning with RLVR}

\author{Hayeon Oh\\
Electronics and Telecommunications Research Institute\\
{\tt\small oph516@etri.re.kr}
}

\begin{document}
\maketitle
\begin{abstract}

Out-of-distribution (OOD) scenarios in autonomous driving pose critical challenges, as planners often fail to generalize beyond their training experience, leading to unsafe or unexpected behavior. Vision-Language Models (VLMs) have shown promise in handling such scenarios by providing high-level scene understanding and user-aligned decisions. However, existing VLMs often exhibit a misalignment between their language-based reasoning and the low-level trajectories required for action-level planning. In this paper, we propose \textbf{LaViPlan}, a framework that leverages Reinforcement Learning with Verifiable Rewards (RLVR) to fine-tune VLMs using planning-oriented metrics. Experimental results show that LaViPlan improves planning performance across both in-domain and out-of-domain datasets. While linguistic fidelity slightly decreases after RLVR-based fine-tuning, qualitative evaluation indicates that the outputs remain coherent. We also conduct ablation studies to analyze the effects of sampling ratio and reasoning guidance, highlighting how these design choices influence performance. These findings demonstrate the potential of RLVR as a post-training paradigm for aligning language-guided reasoning with action-level planning in autonomous driving.

\end{abstract}    
\section{Introduction}
\label{sec:intro}

\begin{figure}[t]
  \centering
  \includegraphics[width=\columnwidth, trim=40 15 20 15, clip]{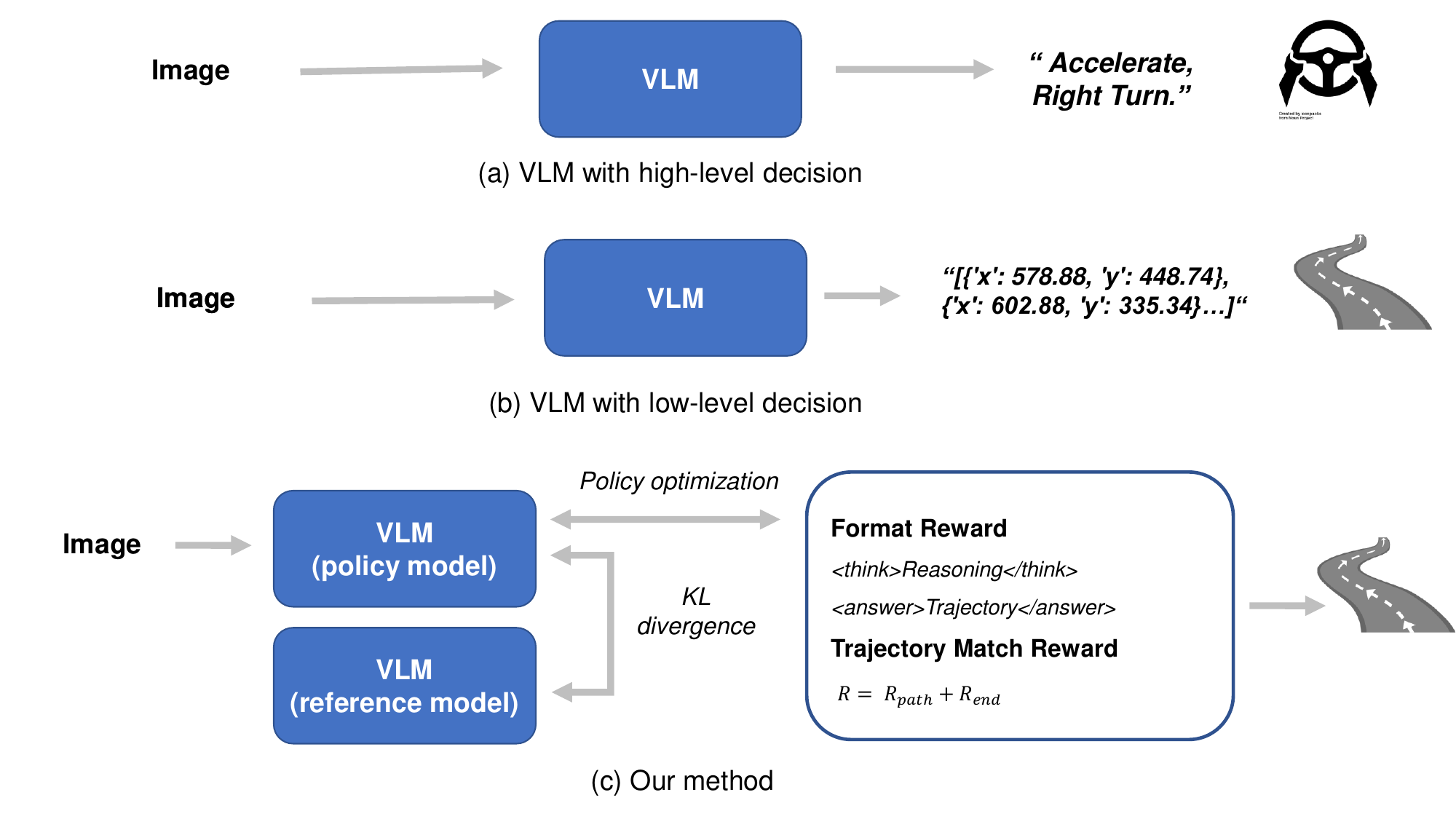}

  \caption{(a) VLMs generating high-level commands (e.g., “Accelerate, Right Turn”) based on scene understanding, but lacking direct trajectory grounding. (b) VLMs producing low-level outputs such as trajectories without explicit reasoning, often leading to semantically inconsistent or context-unaware behavior. (c) Our proposed method introduces a differentiable connection between vision-language reasoning and action space using RLVR. The reward consists of a format-based reasoning verification and trajectory alignment, optimized under KL regularization between the policy and reference model.}
  \label{fig:existing_problems_and_ours}
\end{figure}

Out-of-distribution (OOD) scenarios in autonomous driving refer to rare or novel situations that deviate from the training domain, often leading to unexpected and unsafe behavior by learned planning policies. These scenarios pose critical challenges, particularly when the planner fails to generalize beyond its supervised experience.

To address this, recent research has explored the integration of VLMs \cite{vlm:internvl, vlm:llava, vlm:Qwen-VL} into autonomous driving. VLMs have demonstrated strong generalization capabilities across diverse tasks and modalities, making them a promising approach for handling OOD scenarios. Early research showed that VLMs could identify unseen driving contexts and generate high-level decisions \cite{e2e-vlm:senna, rl:autonomous_alphadrive}.

However, while VLMs can recognize and describe OOD scenes, their final decisions—especially in the form of predicted trajectories—can often be misaligned with the visual reasoning they produce. This issue reflects a broader challenge in aligning language-based reasoning with action-level planning, which we refer to as the \emph{vision-language-action misalignment}.

To mitigate misalignment issues in other domains, GRPO-based reinforcement fine-tuning has been proposed, showing particular effectiveness in improving performance on numerical prediction tasks \cite{rl:visual_rft, rl:manipulator_Maniplvm-r1}. By leveraging multiple candidate outputs and preference-based rewards instead of relying solely on a single ground-truth target, GRPO provides richer supervisory signals that enhance both accuracy and generalization beyond the training distribution. This success suggests that GRPO could be a promising approach to address the misalignment in VLM for autonomous driving.

Building on this insight, we propose to address the misalignment by leveraging Reinforcement Learning with Verifiable Rewards (RLVR), where planning-oriented metrics serve as verifiable reward signals. Our method aims to steer VLMs toward context-aware decision-making that is consistent with their situational reasoning. Our key contributions are summarized as follows:

\begin{itemize}
    \item We propose a reinforcement learning framework that explicitly optimizes planning-oriented metrics in VLMs, demonstrating a step toward aligning language-guided reasoning with action-oriented planning tasks in autonomous driving.
    \item Through both quantitative and qualitative analyses, we reveal that RLVR shifts the model's generation from linguistically faithful outputs to functionally accurate trajectories, indicating a trade-off between semantic similarity and task-specific reasoning.
    \item Experimental results demonstrate that RLVR requires significantly fewer training samples compared to supervised fine-tuning while still achieving performance gains, showing that including hard cases during RL yields better generalization.
\end{itemize}

\section{Related Works}
\label{sec:related_works}

\subsection{VLMs for Autonomous Driving}

End-to-end autonomous driving has recently attracted attention due to its simplicity, efficiency, and ability to mitigate suboptimality arising from misaligned objectives between modular components. By eliminating intermediate representations, this paradigm reduces information loss and computational overhead. However, visual abstraction in end-to-end systems can oversimplify complex scene information, potentially discarding critical cues. Achieving robust generalization across diverse driving scenarios also remains challenging, especially in long-tail scenarios where labeled data is limited or sparse.

To address these limitations, recent studies have explored integrating Vision-Language Models (VLMs) into autonomous driving. VLMs leverage multi-modal understanding to reason about previously unseen situations. Typically, these models combine pre-trained large language models (LLMs) with visual encoders: driving instructions or ego vehicle states are fed as textual input to the LLM, while single- or multi-view images are processed by the visual encoder \cite{e2e-vlm:adapt, e2e-vlm:senna, e2e-vlm:carllava, e2e-vlm:driveVLM, e2e-vlm:lmdrive, e2e-vlm:drivegpt4, e2e-vlm:drivegpt4-v2, e2e-vlm:drivelm}. Recent extensions incorporate 3D perceptual positional embeddings \cite{e2e-vlm:omnidrive, e2e-vlm:orion, e2e-vlm:opendrivevla} and counterfactual learning \cite{e2e-vlm:omnidrive, e2e-vlm:simlingo} to enable context-aware decision-making in complex driving environments.

\subsection{Preference Learning for Alignment}

While VLMs demonstrate strong multi-modal reasoning capabilities, their outputs may still be misaligned with downstream action-level tasks, such as trajectory prediction. One common approach to address this issue is Reinforcement Learning from Human Feedback (RLHF), which aligns VLMs with human preferences using Proximal Policy Optimization (PPO) \cite{rl:ppo, rl:kiminlee-rlhf, rl:autonomous_rlhf}. However, RLHF requires multiple components—including a reference model, reward model, critic, and a newly trained generative model—and relies heavily on human-in-the-loop feedback, resulting in substantial computational and labor costs. This limitation is particularly critical for precise tasks like path planning in autonomous driving.

To overcome these challenges, Group Relative Policy Optimization (GRPO) \cite{rl:deepseek-grpo} has been proposed. GRPO enhances VLMs’ reasoning and arithmetic capabilities by learning directly from comparisons between the policy model and a reference model, without requiring an explicit reward model or critic. The authors advocate for rule-based reward signals, arguing that neural reward models may be vulnerable to reward hacking during large-scale reinforcement learning. Unlike standard or preference-based RL \cite{rl:ppo, rl:dpo}, GRPO directly optimizes verifiable, task-specific metrics. This aligns the policy with downstream objectives, such as path planning, rather than linguistic mimicry, and reduces both human supervision and computational cost.

Building on GRPO, Reinforcement Learning with Verifiable Reward (RLVR) has been successfully applied to tasks where correctness can be objectively evaluated, including object detection, classification \cite{rl:visual_rft}, mathematics \cite{rl:sft_mem_rl_general}, and code generation. RLVR has also been extended to robotics, such as manipulator control \cite{rl:manipulator_Maniplvm-r1, rl:manipulator_world} and autonomous driving \cite{rl:autonomous_waymo_rft, rl:autonomous_alphadrive}, where reward signals can be verified via simulations or pre-defined behavioral criteria. Theoretical analyses show that RLVR with GRPO improves the success rate of reward-maximizing outputs through a dynamic process converging to a fixed point and guarantees preference amplification over the reference model \cite{rl:efficiency_of_grpo_ibm, rl:efficiency_of_grpo_seyoung-yun}.

In this work, we extend RLVR with GRPO to autonomous driving by leveraging planning-oriented metrics, such as Average Displacement Error (ADE) and Final Displacement Error (FDE), as verifiable reward signals. Our approach fine-tunes VLMs to produce trajectories that are consistent with situational reasoning, improving both out-of-distribution generalization and alignment between language-based planning and action-level execution.
\begin{figure*}[t]
\centering
\includegraphics[width=\textwidth]{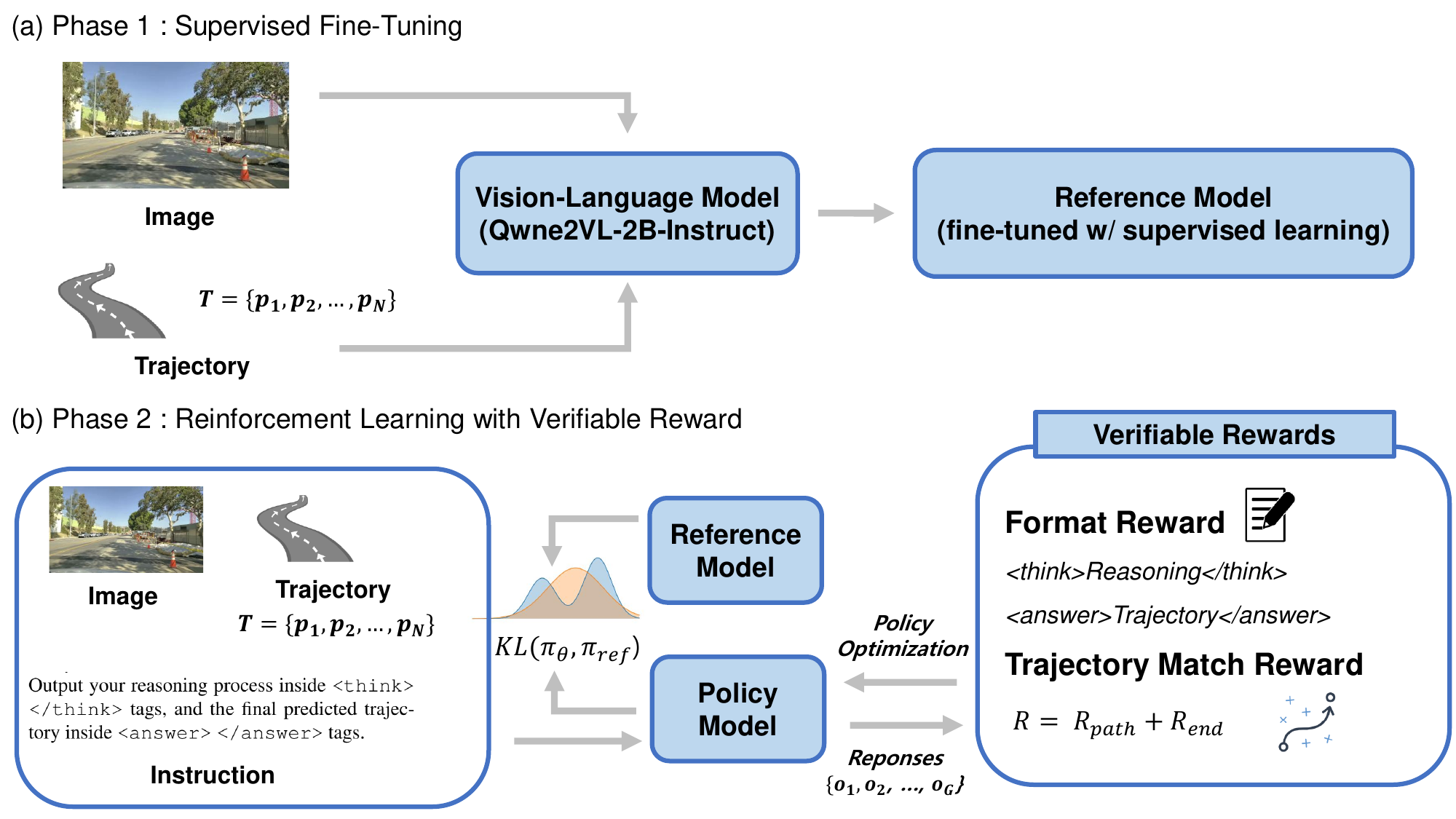}
\caption{Overview of proposed method. (a) In Phase 1, the Vision-Language Model is fine-tuned with supervised learning using paired image-instruction-trajectory data. (b) In Phase 2, reinforcement fine-tuning with verifiable rewards based on format accuracy of responses and trajectory alignment. The policy model is optimized via KL divergence from a supervised reference model with group size of $\mathbf{G}$.}
\label{fig:overview}
\end{figure*}

\section{Methodology}

\subsection{Preliminary} 
\noindent
\textbf{Reinforcement Learning with Verifiable Rewards.} RLVR aims to optimize a policy $\pi_\theta$ by maximizing the expected reward while maintaining proximity to a reference policy $\pi_{\text{ref}}$ to prevent over-optimization. The objective function is formulated as:

\begin{align}
&\max_{\pi_\theta} \; \mathbb{E}_{o \sim \pi_\theta(q)} 
\left[ R_{\text{RLVR}}(q, o) \right] \\ 
&= \left[R(q, o) - \beta \, \mathrm{KL} \left( 
\pi_\theta(o \mid q) \,\|\, 
\pi_{\text{ref}}(o \mid q) 
\right) \right]
\end{align}

where $q$ represents the input query, $o$ denotes the generated trajectory, $R(q, o)$ is the reward function, and $\beta$ is a regularization coefficient that controls the trade-off between reward maximization and KL divergence penalty. The KL divergence term ensures that the learned policy does not deviate significantly from the reference policy, maintaining stability during training.

\noindent
\textbf{Group Relative Policy Optimization.} GRPO extends the Proximal Policy Optimization (PPO)\cite{rl:ppo} by incorporating group-relative advantage estimation. The advantage function is computed using group statistics to reduce variance:

\begin{align}
\hat{A}_{i,t} = \frac{R_i - \text{mean}(\{R_i\}_{i=1}^G)}{\text{std}(\{R_i\}_{i=1}^G)}
\end{align}

where $G$ represents the number of generated trajectories, i.e.\ the group size, $R_i$ is the reward for the $i$-th trajectory, and the advantage $\hat{A}_{i,t}$ is normalized using the mean and standard deviation of rewards within the group. This normalization helps stabilize training by reducing the impact of reward scale variations.

The GRPO objective function combines the clipped surrogate objective from PPO with the group-relative advantages:

\begin{align}
& \mathcal{J}_{GRPO} = \frac{1}{G} \sum_{i=1}^{G} \frac{1}{|o_i|} \sum_{t=1}^{|o_i|} \Bigg[ \min\Bigg\{ \frac{\pi_\theta(o_{i,t}|q, o_{i,<t})}{\pi_{\theta_{\text{old}}}(o_{i,t}|q, o_{i,<t})} \hat{A}_{i,t}, \nonumber \\
&\qquad \text{clip}\left( \frac{\pi_\theta(o_{i,t}|q, o_{i,<t})}{\pi_{\theta_{\text{old}}}(o_{i,t}|q, o_{i,<t})}, 1-\varepsilon, 1+\varepsilon \right) \hat{A}_{i,t} \Bigg\} \nonumber \\
&\qquad - \beta \, \mathrm{KL}[\pi_\theta \| \pi_{\text{ref}}] \Bigg]
\end{align}

where $\pi_{\theta_{\text{old}}}$ represents the policy from the previous iteration, $|o_i|$ is the length of the $i$-th output response, $\varepsilon$ is the clipping parameter, and the min operation implements the conservative policy update mechanism characteristic of PPO. The KL divergence term provides additional regularization to maintain training stability.

\subsection{RLVR with Planning-Oriented Metrics} In this work, we apply GRPO to autonomous driving by designing a reward function that directly reflects planning performance, as shown in \cref{fig:overview}. We build upon the insight that GRPO enables learning pairwise preferences among multiple candidate trajectories. Unlike single-reward updates, this approach exposes the policy to richer and more diverse supervisory signals, providing multiple perspectives on the range of candidate trajectories. This diversity encourages the policy to learn robust behaviors that are not overfitted to specific situations, which ultimately improves generalization across a wide range of scenarios. Therefore, we define the reward using commonly adopted trajectory evaluation metrics: ADE and FDE.

\begin{align}
R_{\text{planning}} ={} 
& -\log \left( 1 + \frac{1}{N} \sum_{i=1}^{N} \left\| \hat{{p}}_i - {p}_i \right\|_2 \right) \notag \\
& -\log \left( 1 + \left\| \hat{{p}}_N - {p}_N \right\|_2 \right)
\end{align}

The first term encourages the predicted trajectory $\hat{T} = \{\hat{p}_1, \hat{p}_2, \ldots, \hat{p}_N\}, \text{ where each } \hat{p}_i \in \mathbb{R}^2$ represents an (x, y) position in image plane to stay close to the ground-truth trajectory ${T} = \{{p}_1, {p}_2, \ldots, {p}_N\}, \text{ where each } {p}_i \in \mathbb{R}^2$  over the full horizon (ADE), while the second term penalizes deviations at the final timestep (FDE). We apply logarithmic smoothing for numerical stability and better learning dynamics. In addition to planning accuracy, we incorporate a formatting reward $R_{\text{format}}$, which encourages adherence to the expected output format with reasoning and response.

\begin{align}
R = R_{\text{format}} + R_{\text{planning}}
\end{align}


\section{Experiment}

\noindent
\textbf{In-Domain Dataset.} The ROADWork dataset \cite{dataset:roadwork} serves as an OOD benchmark focused on road construction scenarios. Among the entire dataset, 5,430 samples contain image, scene description, and a list of trajectories in image plane. To demonstrate that reinforcement fine-tuning can yield performance improvements even with data less than supervised fine-tuning, we divided the dataset into two subsets: 4,344 samples for supervised fine-tuning and 1,086 samples (20\% of full samples) for reinforcement fine-tuning. Details for training can be seen in the Appendix A. 

Furthermore, inspired by KITTI \cite{dataset:kitti}, we split the dataset into two subsets based on the variance of x-coordinates in the trajectory. One subset comprises trajectories with low lateral variance representing straight trajectory \textbf{(Easy)}. In contrast, the other subset comprises trajectories with higher lateral variance and more curved trajectories involving left and right turns \textbf{(Hard)}. Details of the dataset splitting procedure are provided in the Appendix B.

\noindent
\textbf{Out-of-Domain Dataset.}
The CODA-LM \cite{dataset:coda-lm} extends the CODA dataset \cite{dataset:coda} which includes various corner cases such as road construction and adverse weather conditions by augmenting it with natural language captions. The generalization ability of our method under OOD conditions was assessed using the corner-case dataset not limited to construction zones. 

\noindent
\textbf{What is the best baseline model?}
We choose Qwen2VL-2B-Instruct as our baseline. This decision is supported by existing studies \cite{rl:visual_rft, rl:autonomous_alphadrive, rl:limitation_r1_v} that integrate RLVR with diverse tasks and consistently adopt this model, thereby demonstrating its validity and robustness in multimodal learning contexts.

\begin{table}[htbp]
\caption{Instruction for reinforcement fine-tuning and example of response.}
\label{table:dataset_preparation}
\centering
\begin{tcolorbox}[colback=white, colframe=black, boxrule=0.5pt, width=0.95\linewidth]
\textbf{Trajectory Prediction Prompt:} Given only the image, predict the autonomous vehicle's future trajectory as a list of 20 (x, y) image coordinates. You must output your reasoning process inside \texttt{<think>} \texttt{</think>} tags, and the final predicted trajectory inside \texttt{<answer>} \texttt{</answer>} tags. Strictly follow the output format below: \texttt{<think>} Reasoning based on visual cues in the image. \texttt{</think>} \\
\texttt{<answer>[\{'x': x1, 'y': y1\}, ..., \{'x': x20, 'y': y20\}]</answer>}
\end{tcolorbox}

\vspace{5pt}

\begin{tcolorbox}[colback=white, colframe=black, boxrule=0.5pt, width=0.95\linewidth]
\textbf{Example of Response:} \texttt{<think>}tubular markers on right side of road. work vehicle on right side of road. worker on right sidewalk.\texttt{</think>}\texttt{<answer>[\{'x': 578.88, 'y': 448.74\}, \{'x': 578.56, 'y': 442.6\}, ..., \{'x': 602.88, 'y': 335.34\}]</answer>}
\end{tcolorbox}
\end{table}

\noindent
\textbf{Dataset and Prompt for Reinforcement Fine-Tuning.} The dataset for reinforcement fine-tuning consists of a reasoning description enclosed within \texttt{<think></think>} tags, which includes visual reasoning process from the image, and a predicted trajectory represented as a list of image coordinates enclosed within \texttt{<answer></answer>} tags as shown in \cref{table:dataset_preparation}. Specifically, the \texttt{<think></think>} section provides the rationale behind the prediction, while the \texttt{<answer></answer>} section contains the future trajectory as a sequence of coordinate pairs.

\subsection{In-domain Evaluation}

\begin{table}[htbp]
\centering
\small 
\caption{Comparison of model performance on ROADWork dataset. ADE and FDE are presented as percentages by normalizing with image resolution for better readability. The models with supervised fine-tuning were trained on the full dataset of 5K samples, while LaViPlan was fine-tuned with 4K samples for supervised learning and an additional 1K samples for reinforcement fine-tuning. \textbf{Bolded} values indicate the best performance and N/A means no available result.}
\label{table:overview}
\resizebox{1.0\linewidth}{!}{%
\begin{tabular}{l@{\hskip 6pt}cc@{\hskip 6pt}cc}
\toprule
 & \multicolumn{2}{c}{ADE $\downarrow$} & \multicolumn{2}{c}{FDE $\downarrow$} \\
\cmidrule(lr){2-3} \cmidrule(lr){4-5}
& Easy & Hard & Easy & Hard \\
\midrule
\textit{Baseline} &  &  &  &  \\
\multicolumn{5}{l}{\textbf{Vision-Language Models}} \\
Qwen2VL-2B & 52.44 & 52.77 & 102.39 & 105.05 \\
Qwen2VL-7B & 60.73 & 60.71 & 66.61 & 67.57 \\
Qwen2.5-VL-3B & 16.37 & 16.40 & 20.60 & 20.77 \\
LLaMA3.2-11B & 59.27 & 58.88 & 74.16 & 71.44 \\
\multicolumn{5}{l}{\textbf{Domain-Specific Models}} \\
Senna & N/A & N/A & N/A & N/A \\ 
DriveLM (w/ LLaMA-Adapter) & 37.10 & 38.40 & 56.99 & 56.90 \\
\midrule
\textit{Supervised Fine-tuning} & & & & \\
\multicolumn{5}{l}{\textbf{Vision-Language Models}} \\
Qwen2VL-2B & 4.52 & 5.66 & 4.46 & 6.46 \\
Qwen2VL-7B & 4.80 & 6.04 & 5.08 & 7.35 \\
Qwen2.5-VL-3B & 4.97 & 6.22 & 5.07 & 7.34 \\
LLaMA3.2-Vision-11B & 4.52 & 5.46 & 5.20 & 7.10 \\
\multicolumn{5}{l}{\textbf{Domain-Specific Models}} \\
Senna & 5.71 & 5.73 & 6.58 & 7.46 \\ 
DriveLM (w/ LLaMA-Adapter) & 6.73 & 7.79 & 6.87 & 8.43 \\
\midrule
\textit{Reinforcement Fine-tuning} & & & & \\
LaViPlan (ours) & \textbf{3.62} & \textbf{4.83} & \textbf{3.85} & \textbf{6.09} \\
\bottomrule
\end{tabular}%
}
\end{table}

\subsubsection{Overall Results} 
\cref{table:overview} shows overall result across multiple foundation models: baseline, supervised fine-tuning, and our proposed method. They are evaluated with ADE and FDE on both easy and hard subsets. In the baseline, VLMs without additional task-specific supervision lack the capability to generalize directly to trajectory planning. In contrast, supervised fine-tuning improves performance and our proposed method achieves the best performance across all metrics. Compared to supervised fine-tuning models, LaViPlan further reduces the errors. This demonstrates the effectiveness of reward optimization in refining planning-oriented outputs beyond what can be achieved with standard supervised learning. In summary, these results indicate that (1) VLMs are insufficient for planning tasks without supervised fine-tuning, (2) supervised fine-tuning is essential for grounding planning behaviors, and (3) RLVR with planning-oriented metrics leads to further performance gains by explicitly optimizing for planning.

\subsubsection{From Linguistic Consistency to Functional Reasoning} 

\begin{figure}[htbp]
    \centering
    \includegraphics[page=1, width=\linewidth]{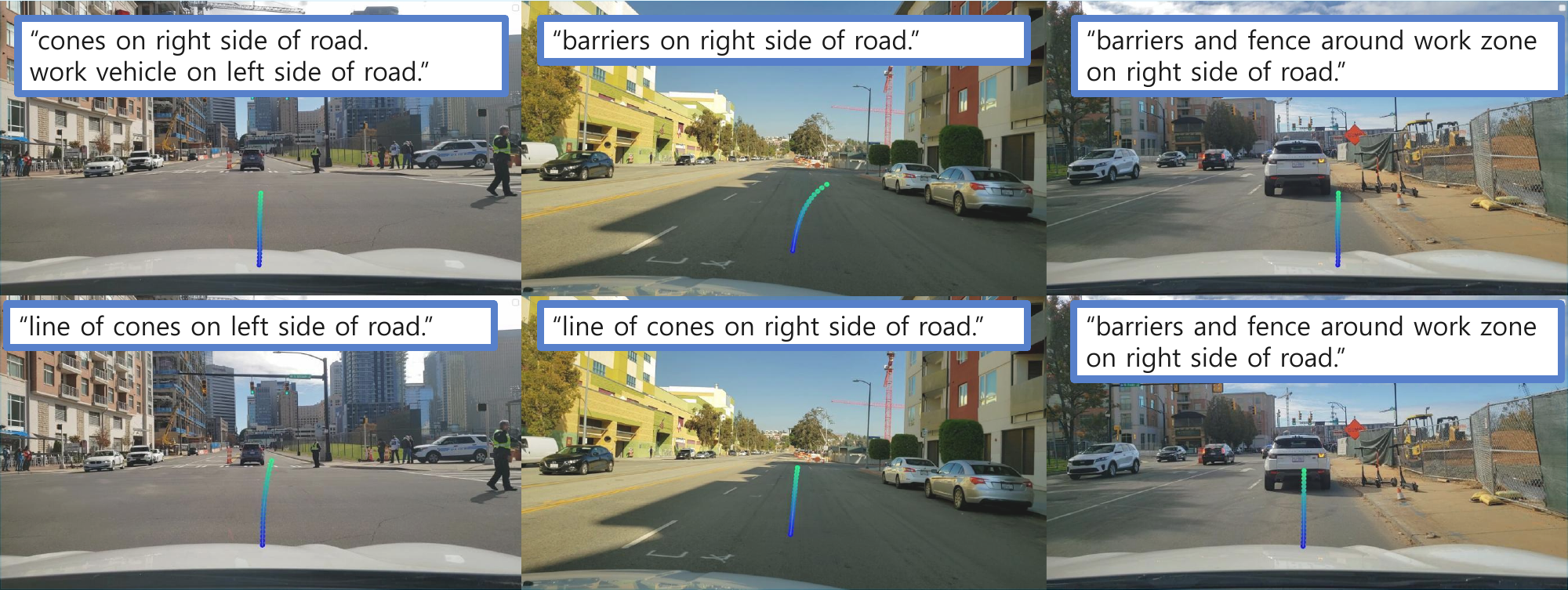}
    \includegraphics[page=2, width=\linewidth]{figures/qualitative_captions.pdf}
    \caption{%
        Qualitative comparison of scene reasoning between the SFT (top) and LaViPlan (bottom) models across six scenarios in \cite{dataset:roadwork}. While SFT tends to produce more verbose, human-like descriptions that closely resemble natural annotations, LaViPlan emphasizes core, task-relevant elements such as cones, barriers, and work vehicles. This shift in reasoning reflects a transition from linguistic fidelity to planning-oriented abstraction, enabling the model to better prioritize hazards and navigable space—even at the cost of similarity to ground-truth descriptions.
    }
    \label{fig:reasoning_shift}
\end{figure}


We evaluate the linguistic effects of our method on the combined easy and hard subsets using BERTScore \cite{metric:bertscore} and Natural Language Inference (NLI) \cite{metric:nli}, as shown in \cref{table:scene_understanding}. Compared to supervised fine-tuning (SFT), reinforcement fine-tuning (LaViPlan) leads to a decrease in BERTScore, as well as a reduction in entailment and an increase in neutral and contradiction labels. While these results indicate a divergence from ground-truth human-written descriptions, they do not necessarily imply degraded reasoning ability. Rather, we observe a meaningful shift in the model’s reasoning style. As shown in \cref{fig:reasoning_shift}, scene descriptions from LaViPlan tend to be more concise and hazard-focused emphasizing critical elements such as barriers, cones, and work vehicles in terms of planning. 

In contrast, SFT tends to replicate verbose and linguistically rich but often functionally redundant information. In several scenarios, LaViPlan omits less relevant descriptors (e.g., sidewalks or drum lines) in favor of spatial cues that better support trajectory-level decisions. This trade-off suggests that conventional language metrics may underrepresent reasoning utility in autonomous driving contexts. Despite lower alignment with human-like phrasing, LaViPlan yields improved planning performance, indicating that more effective and actionable reasoning can emerge even as linguistic similarity declines. We advocate for a shift toward task-aware evaluation of vision-language models in safety-critical domains, where functional relevance should take precedence over semantic mimicry.

\begin{table}[htbp]
\centering
\caption{Comparison of Scene Understanding. SFT refers to supervised fine-tuning and RFT refers to reinforcement fine-tuning}
\label{table:scene_understanding}
\begin{tabular}{lccc}
\toprule
 & \textbf{SFT} & \textbf{LaViPlan} & $\Delta$ \\
\midrule
\textbf{BERTScore} $\uparrow$ & 0.748 & 0.712 & \textcolor{myred}{-4.81\%} \\
\midrule
\textbf{Entailment (\%)} $\uparrow$ & 44.8 & 37.7 & \textcolor{myred}{-15.85\%} \\
\textbf{Neutral (\%)} $\downarrow$ & 25.3 & 29.1 & \textcolor{mygreen}{+15.02\%} \\
\textbf{Contradiction (\%)} $\downarrow$ & 29.9 & 33.1 & \textcolor{mygreen}{+10.70\%} \\

\bottomrule
\end{tabular}
\end{table}

\subsection{Out-of-domain Evaluation} 
We evaluate the proposed method on CODA-LM \cite{dataset:coda-lm}, a zero-shot dataset composed of diverse corner cases involving out-of-distribution (OOD) road hazards. Since CODA-LM does not provide ground-truth trajectories, we evaluate the method based on 2D bounding box in the dataset.

\subsubsection{Evaluation Metrics}

To quantitatively assess the safety of the predicted trajectories, we evaluate three core indicators that capture a distinct aspect of safety. The detailed procedure for obtaining F, C, and P can be found in the Appendix C:

\begin{itemize}
    \item \textbf{Fail Rate (F)}: A trajectory is considered a failure if it invades any 2D bounding box in the scene. This reflects how often unsafe behaviors occur.
    
    \item \textbf{Collision Count (C)}: The average number of bounding boxes that are violated per trajectory across all scenes. This captures the extent of interaction with multiple objects
    
    \item \textbf{Penetration Length (P)}: When a collision occurs, this metric measures the extent of trajectory penetration into the bounding box area, providing a proxy for the severity of the failure. This quantifies the depth of intrusion, emphasizing the potential danger in collision.
\end{itemize}

\noindent 
Inspired by \cite{dataset:bench2drive}, we compute an overall safety score by applying min-max normalization to each metric, followed by a weighted average:

{\footnotesize
\begin{equation}
\text{SafetyScore}_i = \sum_{j \in \{F, C, P\}} w_j \cdot \left(1 - \frac{x_{ij} - \min_k x_{kj}}{\max_k x_{kj} - \min_k x_{kj}} \right)
\label{eq:safety_score}
\end{equation}
}
The safety score for each model $i$ is computed as a weighted sum over metrics $j\in\{F, C, P\}$ — failure rate, collision count, and penetration length — where each metric is min-max normalized across all models $k$. Then, we employ multiple weighting schemes to capture different evaluation perspectives. This approach enables a balanced and comprehensive analysis of model robustness and trade-offs under various out-of-distribution scenarios.

\begin{itemize}
    \item \textbf{Balanced} ($w_f$ = 0.4, $w_c$ = 0.3, $w_p$ = 0.3): This evaluation reflects a balanced trade-off between safety and driving success under general OOD scenarios.

    \item \textbf{Safety-Focused} ($w_f$ = 0.3, $w_c$ = 0.2, $w_p$ = 0.5): Each indicator is first min-max normalized to the [0, 1] range, and then aggregated using predefined weights. This approach highlights the relative performance differences across models and is useful for identifying overall safety trends.

    \item \textbf{Performance-Focused} ($w_f$ = 0.5, $w_c$ = 0.3, $w_p$ = 0.2): This scheme prioritizes driving success by assigning a high weight to the trajectory failure rate, while giving smaller weights to collision count and penetration length, respectively. It emphasizes task completion while still accounting for safety.

    \item \textbf{Equal Weight} ($w_f$ = 0.33, $w_c$ = 0.33, $w_p$ = 0.34): All three safety indicators are weighted equally (approximately 33\% each). This evaluation provides a neutral assessment without emphasizing any specific aspect, making it suitable for overall robustness analysis.
\end{itemize}

\begin{table}[htbp]
\centering
\caption{Safety scores under different weighting schemes for failure (F), collision (C), and penetration (P).}
\label{tab:safety_scores}
\resizebox{1.0\linewidth}{!}{ 
\begin{tabular}{lcccc}
\toprule
\textbf{Model} & \textbf{Balanced}$\uparrow$ & \textbf{Safety-Focused}$\uparrow$ & \textbf{Performance-Focused}$\uparrow$ & \textbf{Equal}$\uparrow$ \\
\midrule
Baseline         & 0.40 & 0.30 & 0.50 & 0.33 \\
SFT (5k)      & 0.60 & 0.59 & \textbf{0.56} & 0.63 \\
LaViPlan    & \textbf{0.64} & \textbf{0.73} & \textbf{0.56} & \textbf{0.70} \\
\bottomrule
\end{tabular} 
}
\end{table}

Table~\ref{tab:safety_scores} presents the safety scores of each model under four different weighting schemes. Across all evaluation settings, the proposed method consistently achieves the highest scores, indicating superior safety performance in both driving success and physical impact mitigation. In particular, it shows a significant advantage under the \textit{Safety-Focused}, minimizing penetration depth and collisions. Although SFT (fine-tuned with whole dataset) performs relatively well, especially under the \textit{Performance-Focused}, it lags behind LaViPlan in overall safety. The baseline model consistently scores the lowest, confirming its limited capability in handling OOD scenarios safely. The qualitative results of SFT and ours can be found in Appendix D.

\subsection{Ablation Studies}

\subsubsection{Supervised vs. Reinforcement Fine-Tuning}
\begin{table}[htbp]
\centering
\caption{Impact of Reinforcement Fine-Tuning (RFT) on performance after Supervised Fine-Tuning (SFT), showing consistent improvements across easy and hard scenarios.}
\label{table:ablation_sft_vs._sft+rft}
\begin{tabular}{lcccc}
\toprule
 & \multicolumn{2}{c}{ADE $\downarrow$} & \multicolumn{2}{c}{FDE $\downarrow$} \\
\cmidrule(lr){2-3} \cmidrule(lr){4-5}
 & Easy & Hard & Easy & Hard \\
\midrule
Baseline & 52.44 & 52.77 & 102.39 & 105.05 \\
SFT (4k) & 4.12 & 5.31 & 4.44 & 6.51 \\
LaViPlan & \textbf{3.62} & \textbf{4.83} & \textbf{3.85} & \textbf{6.09} \\
\midrule
$\Delta$  & 
\textcolor{mygreen}{-12.1\%} & 
\textcolor{mygreen}{-9.1\%} & 
\textcolor{mygreen}{-13.3\%} & 
\textcolor{mygreen}{-6.5\%} \\
\bottomrule
\end{tabular}
\end{table}

We analyze the impact of reinforcement fine-tuning by comparing models with and without reinforcement-based optimization. The observed gains over supervised fine-tuning, as reported in \cref{table:ablation_sft_vs._sft+rft}, indicate that the reinforcement fine-tuning captures complementary learning signals beyond supervised losses. Interestingly, the effect size varies with task complexity, implying that the reinforcement fine-tuning provides stronger gradient signals in low-uncertainty regimes while remaining effective under harder conditions.

\subsubsection{Effective Sample Set for GRPO}

To evaluate the impact of sample difficulty on reinforcement fine-tuning, we fix the SFT:RFT ratio at 4:1 (corresponding to 4,344 and 1,086 samples, respectively), and vary the proportion of easy and hard samples within this fixed budget.  

\noindent
\textbf{In-Domain Dataset.} As shown in \cref{table:ablation_sample_easy_hard}, allocating a larger portion of hard scenarios (up to 40\%) leads to consistent improvements across both ADE and FDE, particularly for hard cases. These results suggest that reinforcement fine-tuning benefits from challenging samples that provide richer learning signals, thereby enhancing the model’s generalization ability in complex planning situations.
Notably, the 6:4 ratio achieves the best overall performance, as it strikes an effective balance between stable learning from easy cases and the additional supervision obtained from hard cases. While increasing the hard portion further (e.g., 7:3) introduces more diverse challenges, it can undermine the stability of learning in easy cases, resulting in less consistent improvements across metrics.

\begin{table}[htbp]
\centering
\scriptsize
\caption{Performance impact of different easy-to-hard data ratio when the total number of training samples is held constant with in-domain dataset.}
\label{table:ablation_sample_easy_hard}
\begin{tabular}{@{}lcccccc@{}}
\toprule
Ratio  & \multicolumn{2}{c}{ADE $\downarrow$} & \multicolumn{2}{c}{FDE $\downarrow$} \\
\cmidrule(lr){2-3} \cmidrule(lr){4-5}
& Easy & Hard & Easy & Hard \\
\midrule
9:1 & 
3.84 \textcolor{mygreen}{(-6.8\%)} & 
5.05 \textcolor{mygreen}{(-4.9\%)} & 
4.09 \textcolor{mygreen}{(-7.9\%)} & 
6.31 \textcolor{mygreen}{(-3.1\%)} \\
7:3 & 
5.55 \textcolor{myred}{(+34.7\%)} & 
6.70 \textcolor{myred}{(+26.2\%)} & 
4.05 \textcolor{mygreen}{(-8.8\%)} & 
6.16 \textcolor{mygreen}{(-5.4\%)} \\
6:4 & 
3.62 \textbf{\textcolor{mygreen}{(-12.1\%)}} & 
4.83 \textbf{\textcolor{mygreen}{(-9.1\%)}} & 
3.85 \textbf{\textcolor{mygreen}{(-13.3\%)}} & 
6.09 \textbf{\textcolor{mygreen}{(-6.5\%)}} \\
\bottomrule
\end{tabular}
\end{table}

\noindent
\textbf{Out-of-Domain Dataset.} The ground-truth trajectory does not always align with the safest path, and such misalignment becomes more prominent in difficult scenarios. In this setting, the 7:3 ratio yields the most favorable outcomes as shown in \cref{table:ablation_sample_easy_hard_ood}, especially under safety-focused and balanced metrics. This is because the larger share of hard samples exposes the model to safety-critical conflicts, encouraging it to learn beyond simply imitating labeled trajectories and to acquire strategies that prioritize safe driving. However, further increasing the proportion of hard samples to 6:4 biases the training distribution excessively toward complex cases, which diminishes generalization in performance-focused metrics.

\begin{table}[H]
\centering
\caption{Performance impact of different easy-to-hard data ratio with out-of-domain dataset and absolute difference from baseline.}
\label{table:ablation_sample_easy_hard_ood}
\resizebox{\linewidth}{!}{ 
\begin{tabular}{lcccc}
\toprule
Ratio / Model & Balanced $\uparrow$ & Safety-Focused $\uparrow$ & Performance-Focused $\uparrow$ & Equal $\uparrow$ \\
\midrule
SFT (5K) & 0.60 \textcolor{mygreen}{(+0.20)} & 0.59 \textcolor{mygreen}{(+0.29)} & 0.56 \textbf{\textcolor{mygreen}{(+0.06)}} & 0.63 \textcolor{mygreen}{(+0.30)} \\ 
LaViPlan (9:1)   & 0.58 \textcolor{mygreen}{(+0.18)} & 0.62 \textcolor{mygreen}{(+0.32)} & 0.51 \textcolor{mygreen}{(+0.01)} & 0.63 \textcolor{mygreen}{(+0.30)} \\
LaViPlan (7:3)   & 0.64 \textbf{\textcolor{mygreen}{(+0.24)}} & 0.73 \textbf{\textcolor{mygreen}{(+0.43)}} & 0.56 \textbf{\textcolor{mygreen}{(+0.06)}} & 0.70 \textbf{\textcolor{mygreen}{(+0.37)}} \\
LaViPlan (6:4)   & 0.45 \textcolor{mygreen}{(+0.05)} & 0.49 \textcolor{mygreen}{(+0.19)} & 0.39 \textcolor{myred}{(-0.11)} & 0.51 \textcolor{mygreen}{(+0.18)} \\
\bottomrule
\end{tabular}
}
\end{table}

In short, our findings indicate that the effectiveness of easy-to-hard sampling is highly domain-sensitive, suggesting that in-domain performance benefits from a balanced exposure to both easy and hard cases, while out-of-domain robustness requires greater emphasis on hard cases to capture safety-critical cues.


\subsubsection{Impact of Reasoning in Two-Phased Fine-tuning}
We ablate the use of explicit reasoning in both supervised fine-tuning (phase 1) and subsequent reinforcement fine-tuning (phase 2) to investigate its effect, as shown in \cref{table:ablation_reasoning_sft} and \cref{table:ablation_reasoning_sft_rft}.

\noindent
\textbf{Reasoning Impact on Supervised Fine-Tuning.} In supervised fine-tuning (phase 1), models trained with reasoning outperformed those without, although the margin was relatively small. Both versions achieved significant gains over the instruction-tuned baseline, with over 90\% reductions in ADE and FDE across all subsets. This indicates that supervised fine-tuning plays a major role in grounding planning behavior, and the addition of reasoning further enhances the model’s ability to contextualize scenes. \\

\begin{table}[htbp]
\centering
\caption{Ablation Study: Effect of Reasoning on Supervised Fine-Tuning (phase 1).}
\label{table:ablation_reasoning_sft}
\resizebox{0.95\linewidth}{!}{ 
\begin{tabular}{lcccc}
\toprule
 & \multicolumn{2}{c}{ADE $\downarrow$} & \multicolumn{2}{c}{FDE $\downarrow$} \\
\cmidrule(lr){2-3} \cmidrule(lr){4-5}
 & Easy & Hard & Easy & Hard \\
\midrule
Baseline & 52.44 & 52.77 & 102.39  & 105.05 \\
w/o reasoning & 
\textcolor{black}{4.14} \textcolor{mygreen}{(-92.10\%)} & 
\textcolor{black}{5.46} \textcolor{mygreen}{(-89.66\%)} & 
\textcolor{black}{4.40} \textbf{\textcolor{mygreen}{(-95.70\%)}} & 
\textcolor{black}{6.65} \textcolor{mygreen}{(-93.67\%)} \\
w/ reasoning & 
\textcolor{black}{4.12} \textbf{\textcolor{mygreen}{(-92.14\%)}} & 
\textcolor{black}{5.31} \textbf{\textcolor{mygreen}{(-89.94\%)}}& 
\textcolor{black}{4.44} \textcolor{mygreen}{(-95.66\%)} & 
\textcolor{black}{6.51} \textbf{\textcolor{mygreen}{(-93.80\%)}} \\
\bottomrule
\end{tabular}
}
\end{table}

\noindent
\textbf{Reasoning Impact on Reinforcement Fine-Tuning.} In reinforcement fine-tuning (phase 2), where reinforcement fine-tuning is applied on top of the already fine-tuned models from phase 1, the inclusion of reasoning continues to offer additional benefits. Although the overall improvements are more modest than those observed in phase 1, models with reasoning consistently outperform those without across both ADE and FDE. The performance gap is slightly larger in the hard subset, suggesting that reasoning becomes more useful in complex scenarios. These findings imply that explicit reasoning, when preserved throughout both fine-tuning stages, provides stable gains by promoting structured understanding aligned with planning objectives.

\begin{table}[htbp]
\centering
\caption{Ablation Study: Effect of Reasoning on Reinforcement Fine-Tuning (phase 2).}
\label{table:ablation_reasoning_sft_rft}
\resizebox{0.95\linewidth}{!}{ 
\begin{tabular}{lcccc}
\toprule
 & \multicolumn{2}{c}{ADE $\downarrow$} & \multicolumn{2}{c}{FDE $\downarrow$} \\
\cmidrule(lr){2-3} \cmidrule(lr){4-5}
 & Easy & Hard & Easy & Hard \\
\midrule
Baseline & 4.12 & 5.31 & 4.44 & 6.51 \\
w/o reasoning & 
\textcolor{black}{3.71} \textcolor{mygreen}{(-9.95\%)} & 
\textcolor{black}{4.95} \textcolor{mygreen}{(-6.78\%)} & 
\textcolor{black}{3.96} \textcolor{mygreen}{(-10.81\%)} & 
\textcolor{black}{6.26} \textcolor{mygreen}{(-3.84\%)} \\
w/ reasoning & 
\textcolor{black}{3.62} \textbf{\textcolor{mygreen}{(-12.14\%)}} & 
\textcolor{black}{4.83} \textbf{\textcolor{mygreen}{(-9.04\%)}} & 
\textcolor{black}{3.85} \textbf{\textcolor{mygreen}{(-13.29\%)}} & 
\textcolor{black}{6.09} \textbf{\textcolor{mygreen}{(-6.45\%)}} \\
\bottomrule
\end{tabular}
}
\end{table}
\section{Discussion}
\label{sec:discussion_limitations}

\noindent
\textbf{Sparse Reward.} RLVR can guarantee performance improvement, but its efficiency may suffer when rewards are sparse. In our setting, the planning reward is sparse, as ADE is only computed after the entire rollout, emphasizing the importance of step-wise feedback during policy optimization. This sparsity may have limited GRPO’s potential, reducing the effectiveness of intermediate decision-making signals \cite{rl:step-wise_GRPO}. Moreover, sparse rewards can slow convergence and increase variance in policy updates, suggesting that integrating denser or auxiliary feedback could further improve learning efficiency and the quality of generated trajectories \cite{rl:limitation_sparse_reward}.

\noindent
\textbf{GRPO as Policy Regularizer, RLVR as Affordance.} GRPO-based fine-tuning is not a silver bullet and relies on the presence of a reasonably strong SFT model as a prerequisite. This is because the policy model is optimized under a divergence constraint that keeps it close to the baseline model. Moreover, the reward function designed for reinforcement fine-tuning defines the affordance of the model, shaping which behaviors are reinforced. Therefore, beyond incorporating existing metrics into the reward design, future work should consider reward formulations that explicitly account for safety, success rate, and other critical factors, rather than solely imitating the ground-truth trajectory.

\section{Conclusion}
In this work, we present LaViPlan, a reinforcement learning framework guided by planning-oriented metrics for fine-tuning VLMs on autonomous driving tasks. Our findings indicate that RLVR enhances the zero-shot scene understanding capabilities of VLMs and may help mitigate misalignment between vision, language, and action. However, the extent to which these benefits generalize to world models—particularly in counterfactual reasoning under actions outside the training distribution in sequential decision-making tasks—remains an open question.

\section{Acknowledgements}
This work was supported by Institute of Information \&
communications Technology Planning \& Evaluation (IITP) grant funded by the Korea government(MSIT) (RS-2023-00236245, Development of Perception/Planning AI SW for Seamless Autonomous Driving in Adverse Weather/Unstructured Environment)

{
    \small
    \bibliographystyle{ieeenat_fullname}
    \bibliography{main}
}

\clearpage
\appendix
\renewcommand\thesection{\Alph{section}}

\section{Training Details}

\begin{table}[htbp]
\centering
\caption{Hyperparameters for language world model training.}
\label{table:hyperparams}
\begin{tabular}{llcc}
\toprule
\textbf{} & \textbf{Hyperparameter} &   \\
\midrule
\multirow{6}{*}{SFT} 
  & Global Batch size        & 128       \\
  & Batch size per GPU       & 4      \\
  & LoRA rank \cite{peft:lora}   & 64          \\
  & LoRA $\alpha$     & 16         \\
  & Epoch             & 1         \\
  & Learning rate     & $1 \times 10^{-5}$\\
  & Weight decay      & 0.1        \\
\midrule
\multirow{6}{*}{RLVR} 
  & Max response length       & 1024         \\
  & Batch size                & 128           \\
  & PPO (GRPO) mini batch size & 4            \\
  & KL loss coefficient       & $0.04$  \\
  & Group size                & 4             \\
  & Learning rate             & $5 \times 10^{-6}$  \\
\midrule
\multirow{2}{*}{Sampling} 
  & Top-$p$        & 0.95          \\
  & Temperature    & 1.2         \\
  & Repetition Penalty    & 1.2         \\
\bottomrule
\end{tabular}
\end{table}

Through an empirical study, we observed that for effective policy model optimization using the GRPO algorithm \cite{rl:deepseek-grpo}, the responses of the reference model must exhibit sufficient diversity within a limited group size. Accordingly, we increased the repetition penalty and temperature slightly compared to the default settings.

\section{Split Dataset}

\subsection{Dataset Construction and Splitting Strategy}

\begin{algorithm}
\caption{Splitting Dataset into SFT and RFT data for Visual Path Planning}
\begin{algorithmic}[1]

\State Load entire trajectory metadata list $D$
\State Compute desired split sizes for:
\begin{itemize}
    \item SFT vs. RFT (e.g., 4:1 ratio)
    \item Straight vs. Turn trajectories (e.g., 6:4 or 4:6 depending on SFT/RFT)
\end{itemize}

\State For each sample $d \in D$, compute variance of $x$-coordinates over its trajectory
\State Sort all samples in descending order of $x$-variance

\State Split into two groups:
\begin{itemize}
    \item Top-$N$ samples with high variance $\rightarrow$ turning samples
    \item Remaining samples $\rightarrow$ straight samples
\end{itemize}

\State From turning samples:
\begin{itemize}
    \item Assign the first $N_1$ to RFT-turn
    \item Assign the remaining $N_2$ to SFT-turn
\end{itemize}

\State From straight samples:
\begin{itemize}
    \item Assign first $M_1$ to SFT-straight
    \item Assign next $M_2$ to RFT-straight
\end{itemize}

\State For each sample in the SFT and RFT sets:
\begin{itemize}
    \item Convert sample format into instruction-following format with:
        \begin{itemize}
            \item An image reference
            \item A natural language prompt
            \item A reasoning + answer pair
        \end{itemize}
    \item Append to respective output list (SFT or RFT)
\end{itemize}

\State Save both lists as JSON files

\end{algorithmic}
\end{algorithm}
To enable both supervised and reinforcement fine-tuning, we split the dataset into Supervised Fine-Tuning (SFT) and Reinforcement Fine-Tuning (RFT) subsets using a structured and variance-aware strategy. We first load the complete set of trajectory metadata $D$ and determine the target ratios for SFT versus RFT (e.g., 4:1), as well as the distribution of turning and straight trajectories within each subset (e.g., 6:4 or 4:6 depending on the setting). For each trajectory $d \in D$, we calculate the variance of its $x$-coordinates, which serves as a proxy for trajectory curvature. We then sort the trajectories in descending order of $x$-variance. Based on this, we classify the top-$N$ trajectories with high variance as turning trajectories and the remaining as straight trajectories. From these, we allocate subsets to SFT and RFT splits: a portion of turning trajectories to RFT-turn and the remainder to SFT-turn, and similarly, straight trajectories to SFT-straight and RFT-straight. Each selected sample is then converted into an instruction-following format consisting of an image reference, a natural language prompt, and a corresponding reasoning-answer pair. 

\subsection{Validation Set Construction}
\vspace{-10pt}  

\begin{figure}[H]
  \centering
  \includegraphics[width=0.8\linewidth]{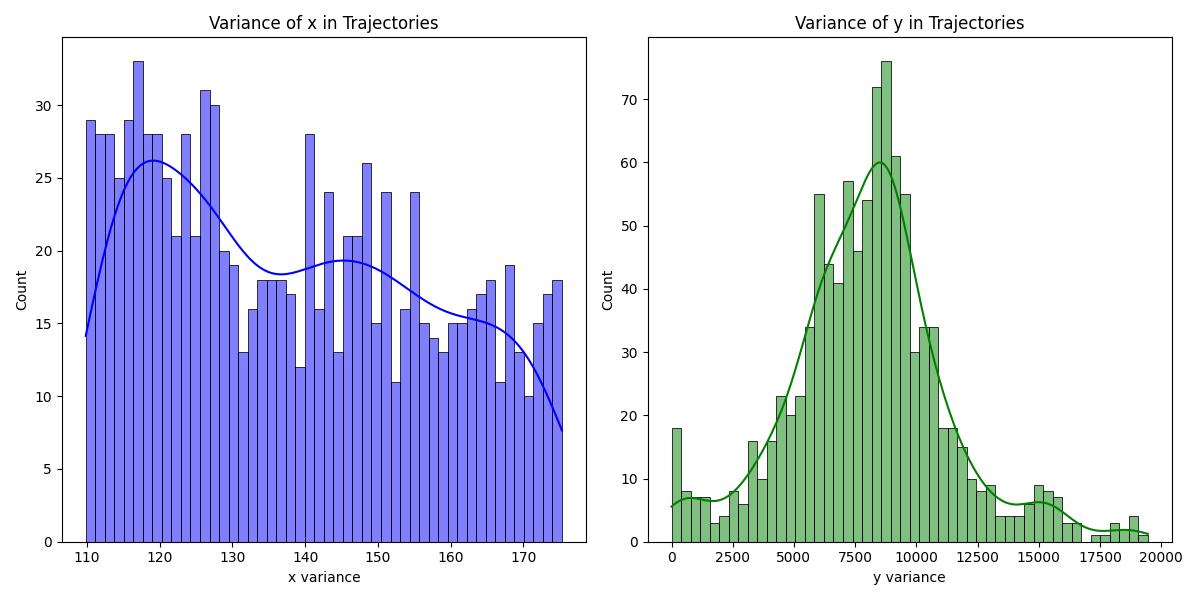}
  \caption{Distribution of trajectories in $D_{easy}$. trajectories exhibit lower x-variance than $D_{hard}$ as can be seen in \cref{fig:var_hard}.}
  \label{fig:var_easy}
\end{figure}
\vspace{-20pt} 
\begin{figure}[H]
  \centering
  \includegraphics[width=0.8\linewidth]{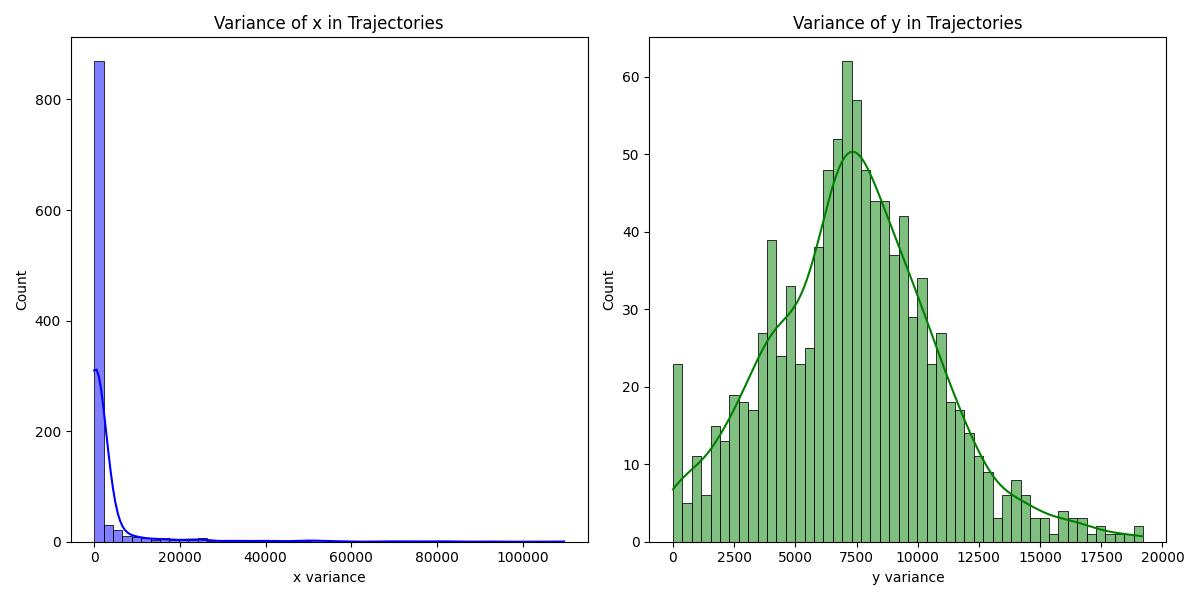}
  \caption{Distribution of trajectories in $D_{hard}$. trajectories exhibit lower x-variance than $D_{easy}$ as can be seen in \cref{fig:var_easy}.}
  \label{fig:var_hard}
\end{figure}

For evaluation, we construct validation sets focused on moderate and hard scenarios as shown in \cref{table:val_split_summary}. We begin with a dense set of validation trajectories $D_{\text{dense}}$ about 12K samples and remove any overlap with standard annotations $D_{\text{standard}}$ to obtain candidate validation samples $D_{\text{val}}$. We compute the $x$-variance for each sample and sort them accordingly. To form the easy set $D_{\text{easy}}$, we select a middle slice (e.g., 1K samples centered around the median $x$-variance). For the hard set $D_{\text{hard}}$, we randomly sample 0.7K trajectories from the top 70\% (high-variance) and 0.3K from the bottom 10\% (low-variance) of the sorted list. 

\setlength{\textfloatsep}{15pt}
\begin{algorithm}[H]
\caption{Construct Validation Sets (Easy, Hard)}
\begin{algorithmic}[1]

\State Load dense validation trajectories $D_{dense}$
\State Load standard validation annotations $D_{standard}$
\State $D_{val} \gets D_{dense} \setminus D_{standard}$
\State For each $d \in D_{val}$, compute variance of $x$ over trajectory
\State $D_{sorted} \gets$ sort $D_{val}$ by $x$-variance (descending)
\State $N \gets |D_{sorted}|$

\State \textbf{Easy Set:} $D_{easy} \gets D_{sorted}[N/2 - 500 : N/2 + 500]$

\State \textbf{Hard Set:} 
\State Randomly sample 700 from top 70\% of $D_{sorted}$
\State Randomly sample 300 from bottom 10\% of $D_{sorted}$
\State $D_{hard} \gets$ combined samples

\State Save $D_{easy}$ and $D_{hard}$ as JSON

\end{algorithmic}
\end{algorithm}

\begin{table}[H]
\centering
\caption{Summary of Validation Subsets Construction}
\label{table:val_split_summary}
\resizebox{\columnwidth}{!}{%
\begin{tabular}{lcc}
\toprule
\textbf{Subset} & \textbf{Sampling Strategy} & \textbf{\#Trajectories} \\
\midrule
Easy ($D_{\text{easy}}$) & Median-centered 1K slice of $x$-variance & 1K \\
Hard ($D_{\text{hard}}$) & 0.7K from top 70\% + 0.3K from bottom 10\% & 1K \\
\bottomrule
\end{tabular}%
}
\end{table}

\section{Pseudo code of OOD Evaluation}
\begin{algorithm}[]
\caption{Procedure for Out-of-Distribution evaluation with as pseudo code.}
\label{alg:coda_evaluation}
\begin{algorithmic}[1]
\Require Dataset $\mathcal{D} = \{(I_i, B_i)\}_{i=1}^N$ where $I_i$: image, $B_i$: 2D bounding boxes
\Require Model $\mathcal{M}$ predicting trajectory $\mathcal{T} = \{(x_j, y_j)\}_{j=1}^{20}$
\Ensure Fail Rate, Avg 2D BBox Collision, Avg Penetration Length

\State \textbf{Initialize:} $C_{fail} \gets 0$, $C_{bbox} \gets 0$, $L_{pen} \gets 0$, $N \gets 0$, $T_{bbox} \gets 0$

\For{each $(I, B)$ in $\mathcal{D}$}
    \State $\mathcal{T} \gets \mathcal{M}(I)$ \Comment{20-point trajectory}
    \State $traj \gets$ LineString($\mathcal{T}$), $c \gets 0$
    
    \For{each $b \in B$}
        \State $p \gets$ Polygon($b$)
        \If{$traj$.intersects($p$)}
            \State $c \gets c + 1$
            \State $L_{pen} \gets L_{pen} + \text{length}(traj \cap p)$
        \EndIf
        \State $T_{bbox} \gets T_{bbox} + 1$
    \EndFor

    \If{$c > 0$} $C_{fail} \gets C_{fail} + 1$ \EndIf
    \State $C_{bbox} \gets C_{bbox} + c$, $N \gets N + 1$
\EndFor

\State \Return 
\quad $ \frac{C_{fail}}{N}$,\quad
$\frac{C_{bbox}}{N}$,\quad
$\frac{L_{pen}}{N}$
\end{algorithmic}
\end{algorithm}

\textbf{Trajectory Fail Rate:} The proportion of predicted trajectories that intersect with at least one 2D bounding box in the scene:
\begin{equation}
\text{Fail Rate} = \frac{\text{Number of collision trajectories}}{\text{Total number of samples}}
\end{equation}

\noindent 
\textbf{Average 2D BBox Collision:} The average number of bounding boxes that each predicted trajectory collides with:
\begin{equation}
\text{Collision Count} = \frac{\text{Total bbox collisions}}{\text{Total number of samples}}
\end{equation}

\noindent
\textbf{Penetration Length:} The average geometric length of trajectory segments that penetrate into bounding boxes:
\begin{equation}
\text{Avg Penetration Length} = \frac{\sum_{i=1}^{N} \sum_{j=1}^{|B_i|} \text{length}(\mathcal{T}_i \cap bbox_j)}{\text{Total number of samples}}
\end{equation}

where $\mathcal{T}_i \cap bbox_j$ represents the intersection between trajectory $\mathcal{T}_i$ and bounding box $bbox_j$.

\section{Qualitative Results on OOD Benchmark}

\begin{figure*}[]
    \centering
    \begin{subfigure}{\textwidth}
        \centering
        \includegraphics[width=\textwidth]{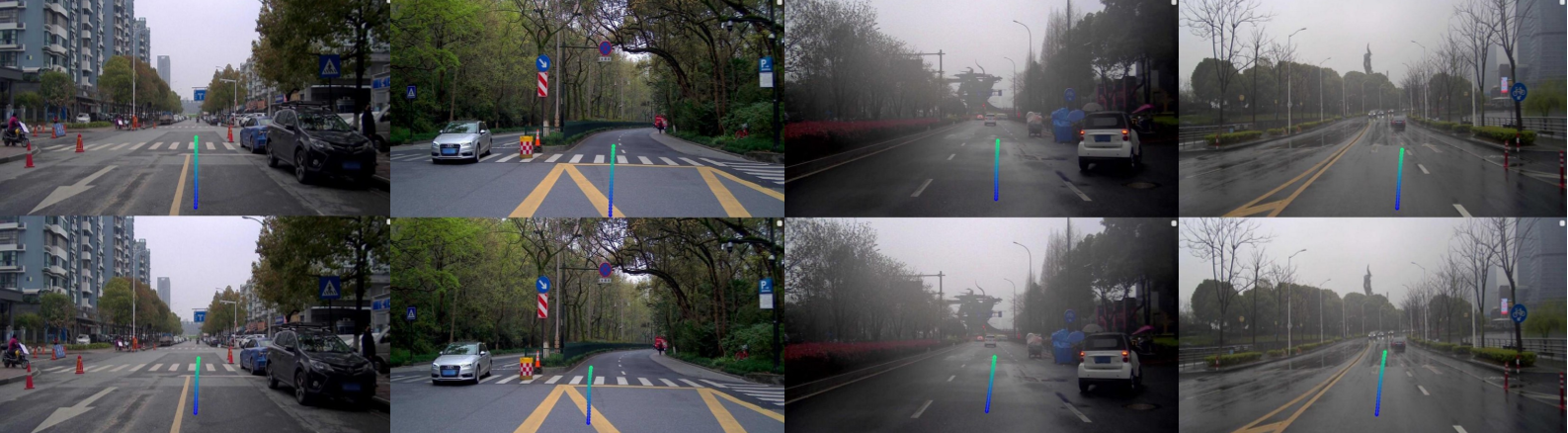}
    \end{subfigure}
    \vspace{0.1em}  
    \begin{subfigure}{\textwidth}
        \centering
        \includegraphics[width=\textwidth]{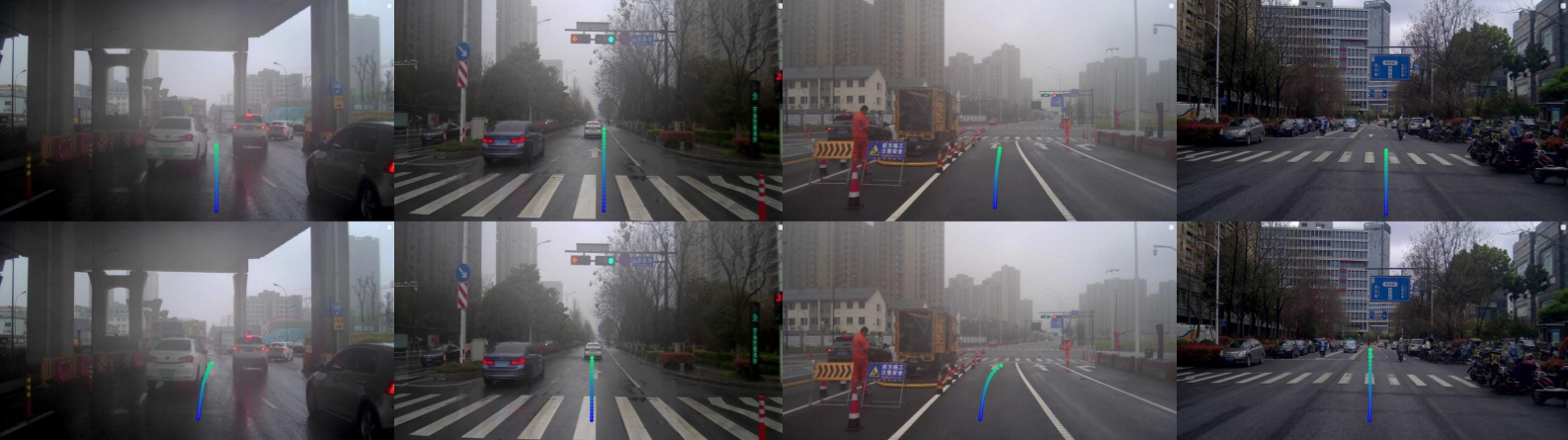}
    \end{subfigure}
    \vspace{0.5em}
    \begin{subfigure}{\textwidth}
        \centering
        \includegraphics[width=\textwidth]{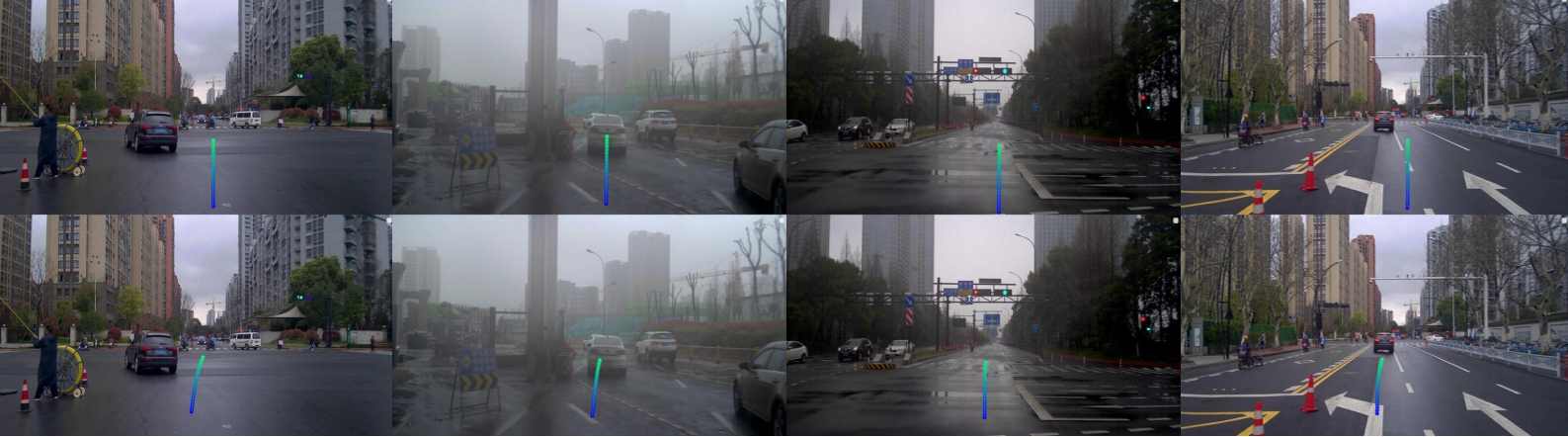}
    \end{subfigure}

    \caption{Qualitative results comparing supervised fine-tuning (top) and RLVR-based reinforcement fine-tuning (bottom) across diverse out-of-distribution (OOD) hazard scenarios. Examples include construction zones, adverse weather (e.g., rain and poor lighting), unexpected pedestrian behavior, and road obstacles. The proposed method generates more accurate and context-aware trajectories under complex conditions, indicating better robustness in real-world hazard cases.}
    \label{fig:qualitative_output}
\end{figure*}

We qualitatively evaluate the proposed method on CODA-LM, a zero-shot dataset involving diverse corner cases and road hazards. Since CODA-LM does not provide ground-truth trajectories, we rely on visual inspection of the predicted paths to assess plausibility, safety, and contextual awareness. As shown in \cref{fig:qualitative_output}, the model fine-tuned with verifiable rewards produces more realistic and contextually appropriate trajectories compared to the one trained via supervised fine-tuning. In scenarios involving roadwork, blocked lanes, and ambiguous path constraints, the supervised model often generates linear or risk-prone trajectories that lack appropriate deviation or caution. In contrast, the proposed method consistently generates smoother, safer, and more context-aware trajectories, often adjusting path curvature to avoid obstacles such as cones, barriers, and vehicles. These results demonstrate the model’s capacity to generalize to previously unseen situations by aligning its reasoning and path generation with latent planning cues in the scene, even without explicit supervision. This qualitative result suggests that reinforcement fine-tuning enhances the model’s ability to adapt to driving contexts.

\end{document}